\documentclass[graybox,envcountchap,sectrefs,a4paper]{svmono}

\usepackage{type1cm}         
\usepackage{makeidx}         
\usepackage{graphicx}        
\usepackage{multicol}        
\usepackage[bottom]{footmisc}

\usepackage{newtxtext}       %
\usepackage{newtxmath}       

\usepackage{algorithm}
\usepackage{algorithmic}
\usepackage{float}
\usepackage{xcolor}
\usepackage{svg}
\usepackage{caption}
\usepackage{subcaption}
\usepackage{tikz}
\usepackage{amsmath}
\usepackage{wrapfig}
\usepackage{gensymb}

\newcommand{\highlight}[1]{{\color{red!100!black!100}#1}}

\begin{document}
\makeatletter
\newcommand{\chapterauthor}[1]{%
  {\parindent0pt\vspace*{-110pt}%
  \linespread{1.1}\large\scshape#1%
  \par\nobreak\vspace*{15pt}}
  \@afterheading%
}
\makeatother

\chapter{High-bandwidth nonlinear control for soft actuators with recursive network models}

\chapterauthor{Sarah Aguasvivas Manzano\footnote{Department of Computer Science, University of Colorado Boulder, Boulder, CO, 80309,
USA}, Patricia Xu\footnote{Sibley School of Mechanical and Aerospace Engineering, Cornell University, Ithaca,
NY 14850, USA}, Khoi Ly\footnote{Paul M. Rady Department of Mechanical Engineering, University of Colorado Boulder,
Boulder, CO 80309, USA}, Robert Shepherd$^2$ and Nikolaus Correll$^1$}

\abstract{We present a high-bandwidth, lightweight, and nonlinear output tracking technique for soft actuators that combines parsimonious recursive layers for forward output predictions and online optimization using Newton-Raphson. This technique allows for reduced model sizes and increased control loop frequencies when compared with conventional RNN models. Experimental results of this controller prototype on a single soft actuator with soft positional sensors indicate effective tracking of referenced spatial trajectories and rejection of mechanical and electromagnetic disturbances. These are evidenced by root mean squared path tracking errors (RMSE) of $1.8mm$ using a fully connected (FC) substructure, $1.62mm$ using a gated recurrent unit (GRU) and $2.11mm$ using a long short term memory (LSTM) unit, all averaged over three tasks. Among these models, the highest flash memory requirement is $2.22kB$ enabling co-location of controller and actuator.}

\section{Introduction}
\label{sec:1}

\begin{wrapfigure}{r}{0.45\columnwidth}
    \vspace{-60pt}
    \centering
    \begin{tikzpicture}
    \node (image) at (0, 0) {\includegraphics[width=0.45\columnwidth, trim={10cm 0cm 0cm 3cm},clip]{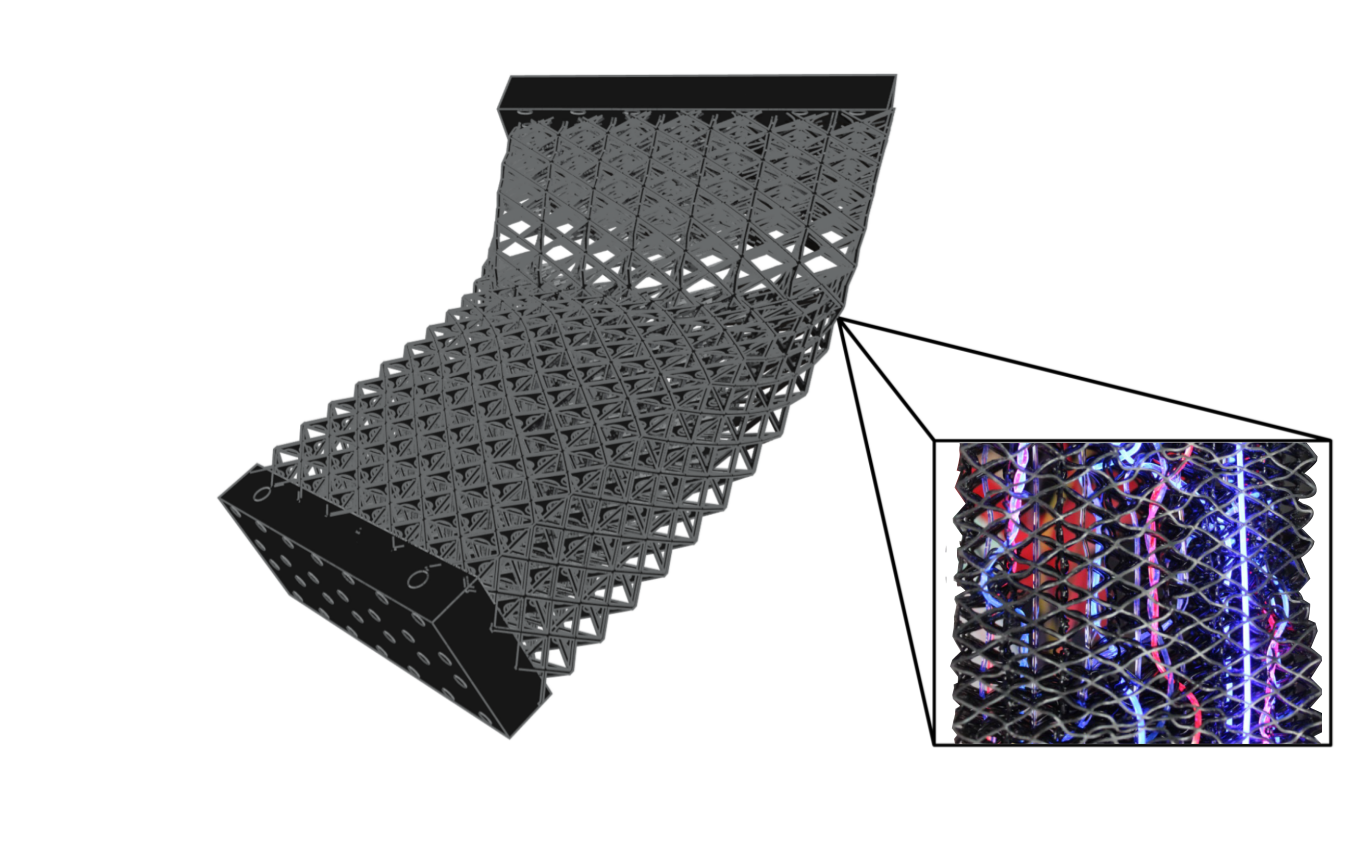}};
    \end{tikzpicture}
    \vspace{-18pt}
    \caption{Rendering of the mesh with embedded light lace sensors.}
    \vspace{-16pt}
    \label{fig:fig1}
\end{wrapfigure}

Model-free, learning based approaches to control soft devices have proven to outperform model-based approaches for real-life systems \cite{cecilia-laschi}. In most robotic designs, the placement and design of embedded sensors not only determines the sensing resolution and reliability, but also how the sensors are calibrated and modeled \cite{ml4soft}. In the case of soft sensors, complications emerge due to the shifts in position caused by time dependent mechanical changes such as hysteresis, creep, and fatigue. The system used in this work is a soft artificial muscle  embedded with a soft optical sensor network that we call ``Light Lace'' \cite{optical} described in Fig. \ref{fig:fig1}.

\runinhead{Related Work.} State-of-the-art approaches found in \cite{ml4soft} address the challenges in machine learning (ML) for modeling and control of soft actuators. In \cite{hyatt}, a very large ($3.4$ million nodes) neural network model that learns the forward dynamics of the system is used for model predictive control. This model is converted to a linear state space model where the system and input matrices are extracted from a neural network model discretized at $0.05s$, thus controlled at $20 Hz$. In \cite{gillepsie}, a fully connected neural network of three hidden layers with $200$ nodes each represents a discretized model at $0.033s$. For the control decision, the solver CVXGEN \cite{cvxgen} requires a linearized model coming from automatic differentiation from a high-level deep learning package. The control loop frequency ($30Hz$) is limited by the sampling rate of the sensor used in this application. Lastly, in \cite{bruder_gillepsie} authors achieved path tracking errors in the order of centimeters (average $L_2$ error of $1.26cm$) and had noisy output tracking when the robot was performing tasks. Finally,  \cite{rnn_thurtel} has shown that Long Short-Term Memory units (LSTM), are suitable for modeling the kinematic responses of soft actuators with embedded sensors in real time while being robust against sensor drift.

The aforementioned results may be limited by either the sampling rate of the system, a very large neural network model to represent the forward kinematics, or a limiting experimental mechanism that does not allow for smooth path tracking on the actuator. Complex models with large memory requirements are often difficult to embed into resource-limited microcontrollers \cite{nn4mc}
without any model reduction technique or fixed point approximation. Our approach seeks to simplify the model representation needed to control a soft actuator.  This would allow to create materials that deeply embed sensing, computation and actuation, thereby leading to ``materials that make robots smart'' \cite{hughes2019materials}. Yet, previous literature does not provide insights on techniques for automatic differentiation in platforms that cannot use high-level neural network packages enabled with automatic differentiation. 

\runinhead{Contribution of this paper}

We present a nonlinear, predictive controller capable of functioning at high bandwidth through the use of parsimonious recursive networks and online optimization using a Newton-Raphson solver \cite{ngpc}. We develop this controller prototype and software architecture that are tractable in an off-the-shelf microcontroller, thus we aim for low space complexity and low memory footprint. Our approach compares favorably with similar state-of-the-art approaches \cite{hyatt, gillepsie, bruder_gillepsie} by decreased latency through reproducible numerical gradient approximations, decreased output tracking errors, increased real-time smoothness, proven repeatability and precision experimentally despite sensor input being highly nonlinear. An implementation of our framework is available open-source. \footnote{https://github.com/sarahaguasvivas/nlsoft} Finally, we demonstrate that large neural network models are not necessary for control and identification of soft robotic actuators with a low number of states to achieve an accuracy competitive with state-of-the-art approaches. 

\section{Experimental Setup}
\label{sec:experimental_setup}

\runinhead{Problem Statement}
\label{sec:2}

We consider a segment of a fully flexible, soft mesh mounted in a solid structure, where the end effector's pose is defined as $\mathbf{y} = \{x_0, ... x_{n-1} \}$ and is actuated by $m$ different tendon actuators $\mathbf{u} = \{u_0, ..., u_{m-1} \}$. Embedded in this mesh are $w$ channels of light lace sensors \cite{optical} denoted as $\mathbf{l} = \{l_0, l_1, ..., l_{w-1} \}$. %

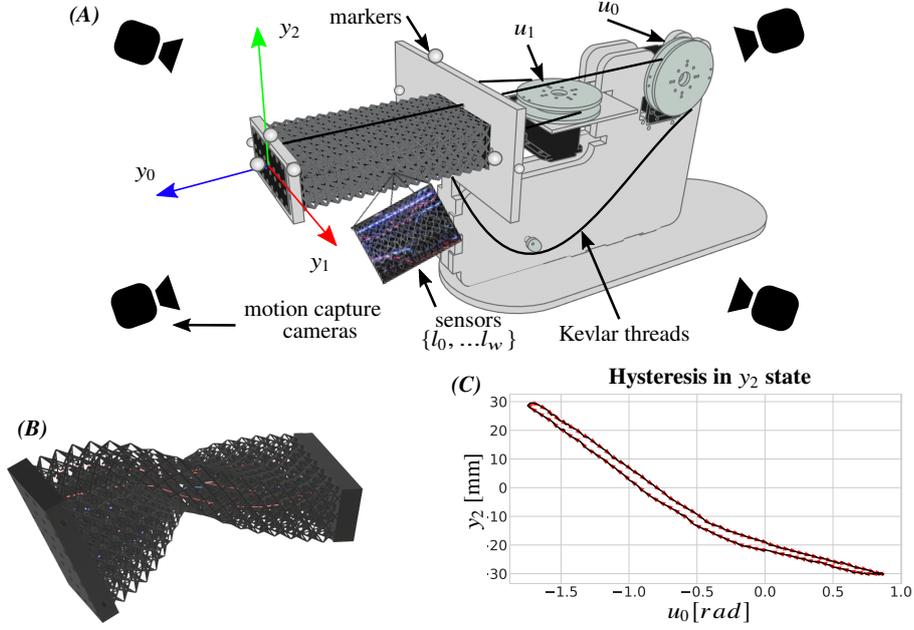
\begin{figure}[H]
    \vspace{-15pt}
    \centering
      \begin{subfigure}[t]{\textwidth}
    \centering
    \begin{tikzpicture}
    \begin{scope}
    \node (image) at (0, 0)
    {\includegraphics[height=0.4\linewidth]{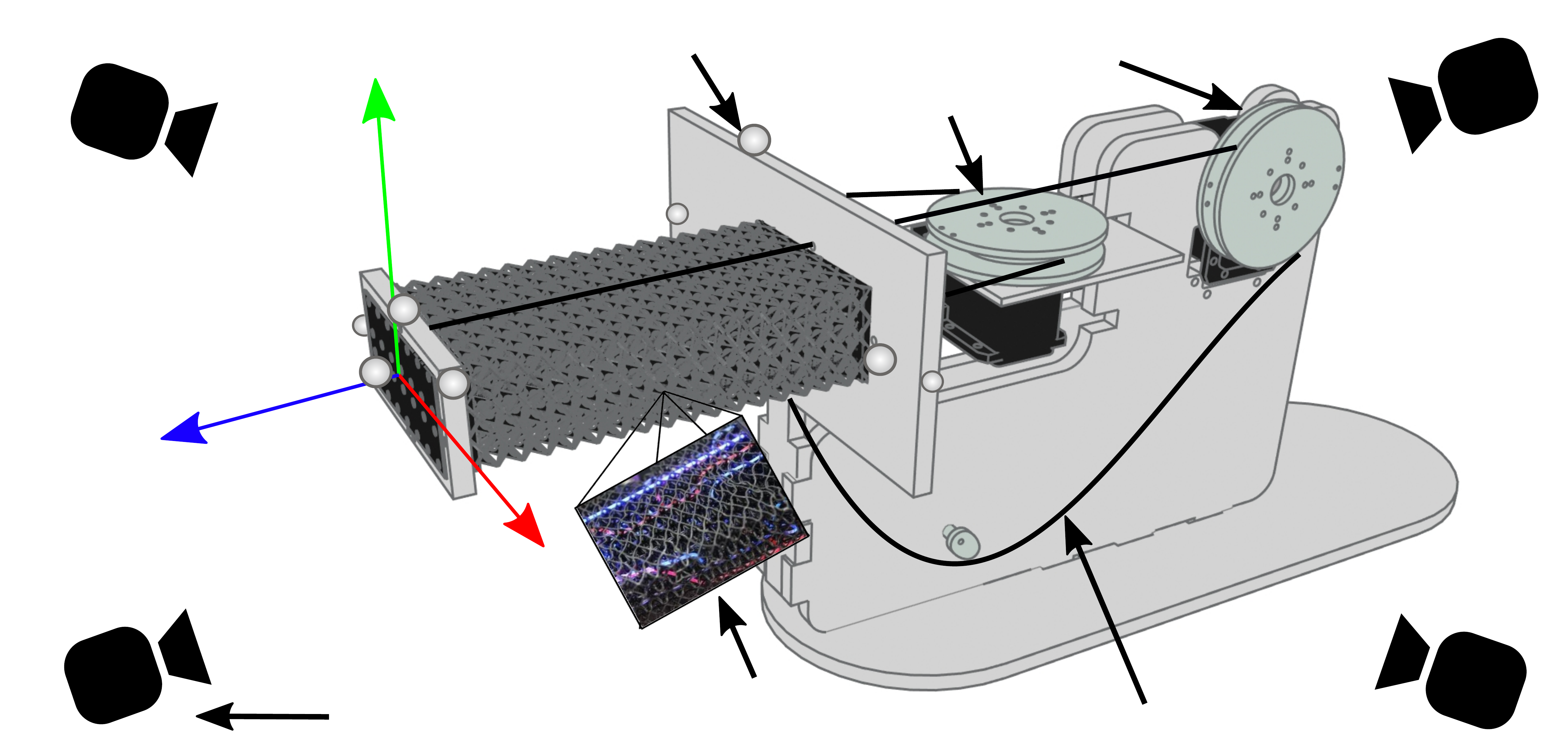}};
    \end{scope}
    \node (note) at (-2.1,1.8) {$y_2$};
    \node (note) at (-4,-0.1) {$y_0$};
    \node (note) at (-1.7,-1.3) {$y_1$};
    \node (note) at (-1.1,2.0) {markers};
    \node (note) at (-1.8,-1.9) {motion capture};
    \node (note) at (0.25,-2.05) {sensors};
     \node (note) at (0.25,-2.3) {$\{ l_0, ... l_w \}$};
    \node (note) at (-1.7,-2.2) {cameras};
    \node (note) at (1.0,1.8) {$u_1$};
    \node (note) at (2.1,2.1) {$u_0$};
    \node (note) at (2.3,-2.2) {Kevlar threads};
    \node (note) at (-4.8,2) {\textit{\textbf{(A)}}};
    \end{tikzpicture}
    \end{subfigure}
    
    \begin{subfigure}[t]{0.4\textwidth}
        \centering
        \begin{tikzpicture}
        \begin{scope}
        \node (image) at (0, 0) {\includegraphics[width=1\linewidth]{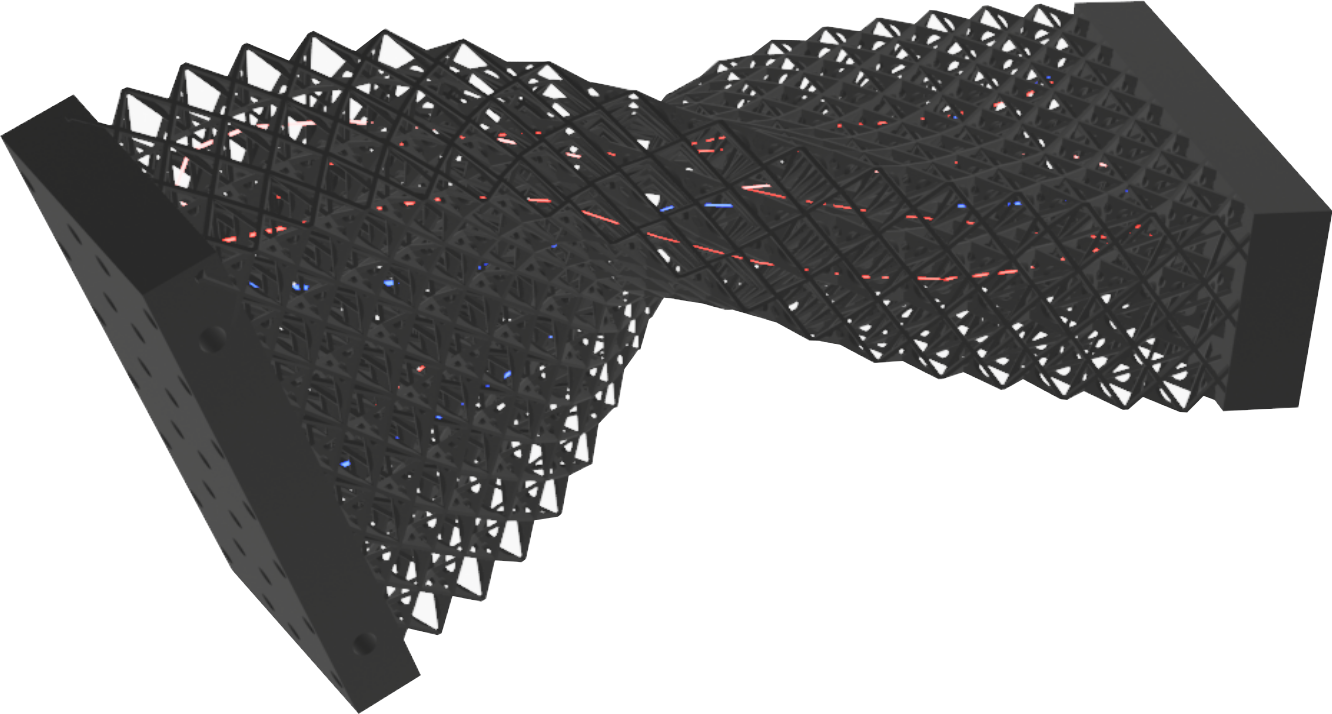}};
        \end{scope}
         \node (note) at (-2,1.3) {\textit{\textbf{(B)}}};
        \end{tikzpicture}

    \end{subfigure}
    \hfill
    \begin{subfigure}[t]{0.5\columnwidth}
        \centering
        \begin{tikzpicture}
        \node (image) at (0, 0) {\includegraphics[width=\columnwidth, trim={2.05cm 0.8cm 1cm 1cm},clip]{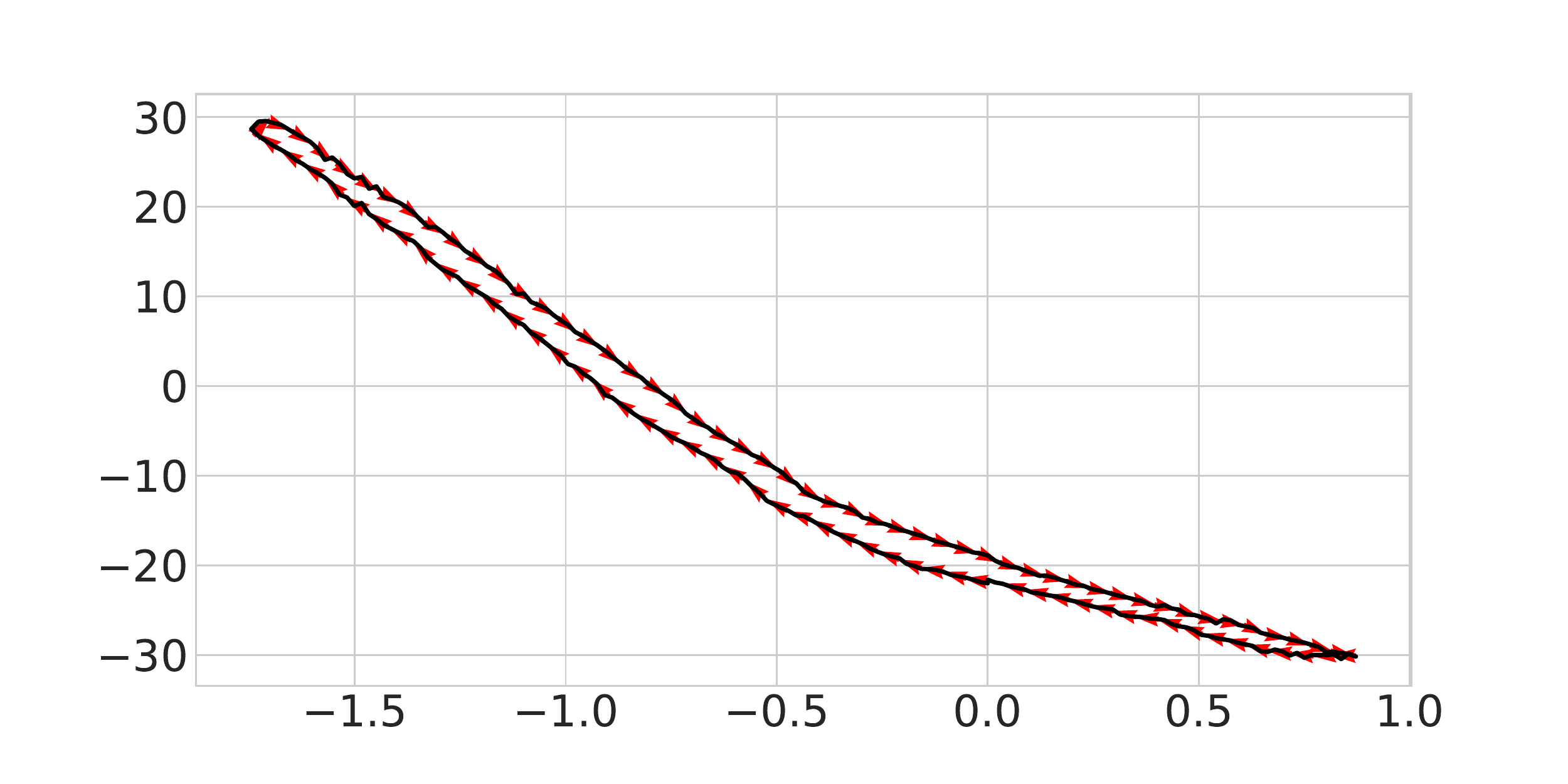}};
        \node (note) at (0,1.5) {\textbf{Hysteresis in $y_2$ state}};
        \node (note) at (-3.2,1.4) {\textit{\textbf{(C)}}};
        \node[rotate=90] (note) at (-3.1,0.0) {$y_2$ [mm]};
         \node (note) at (0,-1.6) {$u_0 [rad]$};
        \end{tikzpicture}
    \end{subfigure}
    \caption{Problem statement. \textit{(A)} Experimental setup and variable definitions. \textit{(B)}  Artistic rendering to show the flexibility of the mesh and sensor materials in \textit{(A)}. \textit{(C)}  Non-linearity and hysteresis with respect to control input.}
    \label{fig:problem}
    \vspace{-10pt}
\end{figure}

\runinhead{Materials and Methods} 
Fig. \ref{fig:problem} \textit{(A)} describes the experimental test bed in this work. The undeformed mesh dimensions are $14\times 7 \times 3 cm^3$ in the $y_0, y_1, \text{ and } y_2$ dimensions respectively. Two high torque servos (Dynamixel RX-64, Robotis) pull tendons that move the lattice's end effector in 3-dimensional coordinates. This end effector can move up to $\Delta \mathbf{y}_{range} = \{40.4 ,28.3, 35.0\} mm $ in each direction. The tendons are connected to the end effector using high-strength Kevlar threads (Kevlar Fiber, Dupont). The inputs indicate the angular displacements of the servos with respect to the manufacturer reference datum. The optical lace sensor network consists of 11 normalized channels that are organically weaved within the soft mesh structure. We use a motion capture system (Optitrack, Natural Point, Inc.) to track the relative spatial position of the end effector with respect to its mounting base. In the real-time computation of the optimal control output, the motion capture system signals are not used towards the controls computations, but we record true positions of the end effector for evaluation in our results and computation of the errors in Sec. \ref{sec:results}. 

\textit{Light lace sensor network.} Our system is composed of a tendon-driven, soft, 3D printed, polyurethane mesh with stretchable lightguides distributed within the body. The $1mm$ diameter, polyurethane fibers (Crystal Tec) intertwine to make up a network of optical lace sensors \cite{optical} consisting of input lines that carry light from LEDs and output lines that carry coupled light to photodiodes. The coupled light intensities vary based on the contact between input and output fibers as the system deforms to provide local strain information within the soft mesh to differentiate between twisting, bending and stretching states. This sensor network stretches and warps together with the soft mesh as seen in Fig. \ref{fig:problem} \textit{(B)}.

\vspace{-10pt}

\section{Nonlinear Online Controller}
\label{subsec:4}

Similar to \cite{ngpc, rnn_thurtel}, we formulate a discretized output transition model that uses the feedback from the sensors and a history of past inputs and output estimates. We describe this model as $\dot{\mathbf{y}}(t) = g(\mathbf{\tau}, \mathbf{\alpha}, \mathbf{l})$, where $\mathbf{\tau}$ is a queue composed by $\{\mathbf{u}(t-n_d), ... ,\mathbf{u}(t)\}$;  $\mathbf{\alpha}$ is composed by $\{\mathbf{y}(t-d_d), ... ,\mathbf{y}(t-1)\}$ and $d_d$ and $n_d$, which are how far into the past the model looks at predictions and inputs respectively. We define $N$ as the prediction horizon, that is, the number of times the nonlinear discrete model will be recursively called in order to predict future outputs with $N_1$ and $N_2$, ($N_1 < N_2 \leq N$) being the start and end of the cost horizon defined in Eq. \ref{eq:cost}. $N_c (\leq N )$ is the control horizon, which is used towards the prediction and when $N_c < N$, we roll the $\tau$ queue until $N_c$ and then repeat $\mathbf{u}_{N_c}$ until $N$.

\runinhead{Step 1: Data collection} 
\label{subsec:5}

We record time series data from the sensor signals, the motion capture system, and the servo inputs $\mathbf{u}$. We run a three-stage sequence of inputs illustrated in the first row of Fig. \ref{fig:model}, where samples of these sequences are shown along with the learned model offline predictions. Samples of this input sequence at each stage is described in the first row of Fig. \ref{fig:model} and is repeated 200 times. We then prepare the data depending on the values of $n_d$ and $d_d$. This results in a data size of $3.8$ million samples, which are then reorganized depending on $n_d$ and $d_d$. The signals were collected at a discretization of an average step duration of $8.\bar{3} ms$ or $120 Hz$ for a total of $8.8$ hours. 

\runinhead{Step 2: Learning the model}
\label{sec:model}

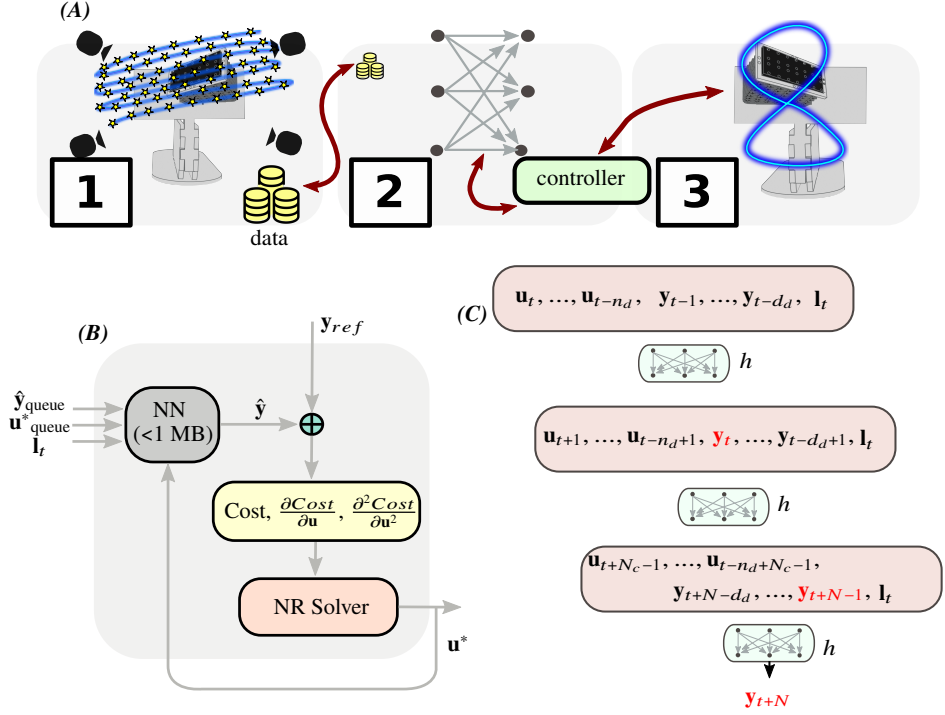
\begin{figure}[H]
    \vspace{-15pt}
    \centering
     \begin{subfigure}[t]{\textwidth}
    \centering
    \begin{tikzpicture}
    \begin{scope}
    \node (image) at (0, 0)
    {\includegraphics[width=1\columnwidth]{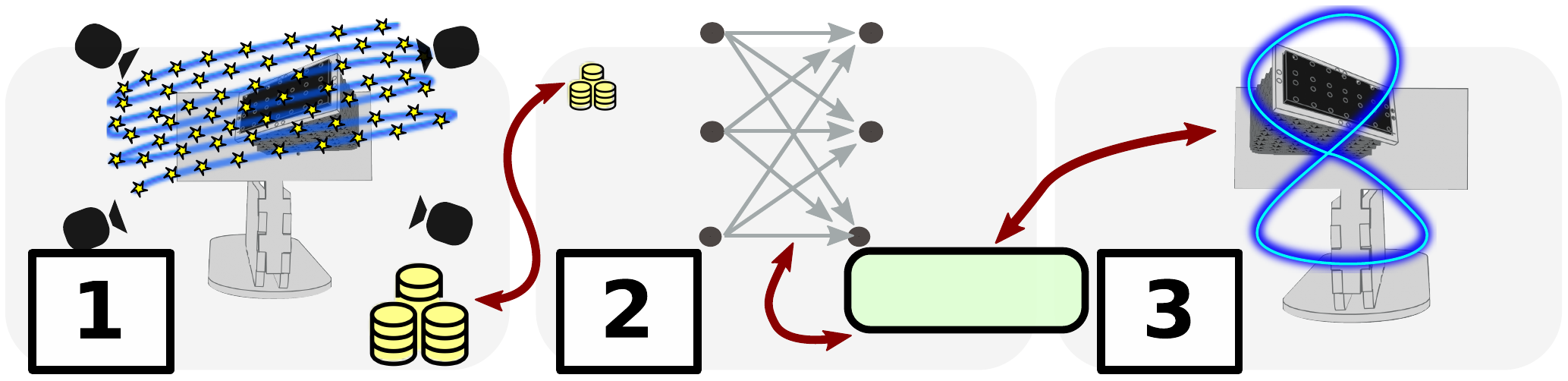}};
    \end{scope}
    \node (note) at (1.35,-0.7) {controller};
    \node (note) at (-2.75,-1.5) {data};
    \node (note) at (-5.3,1.5) {\textit{\textbf{(A)}}};
    \end{tikzpicture}
    \end{subfigure}
    
    \begin{subfigure}[t]{0.5\columnwidth}
    \hspace{-25pt}
        \centering
        \begin{tikzpicture}
        \begin{scope}
        \node (image) at (0,0)
        {\includegraphics[width=\columnwidth]{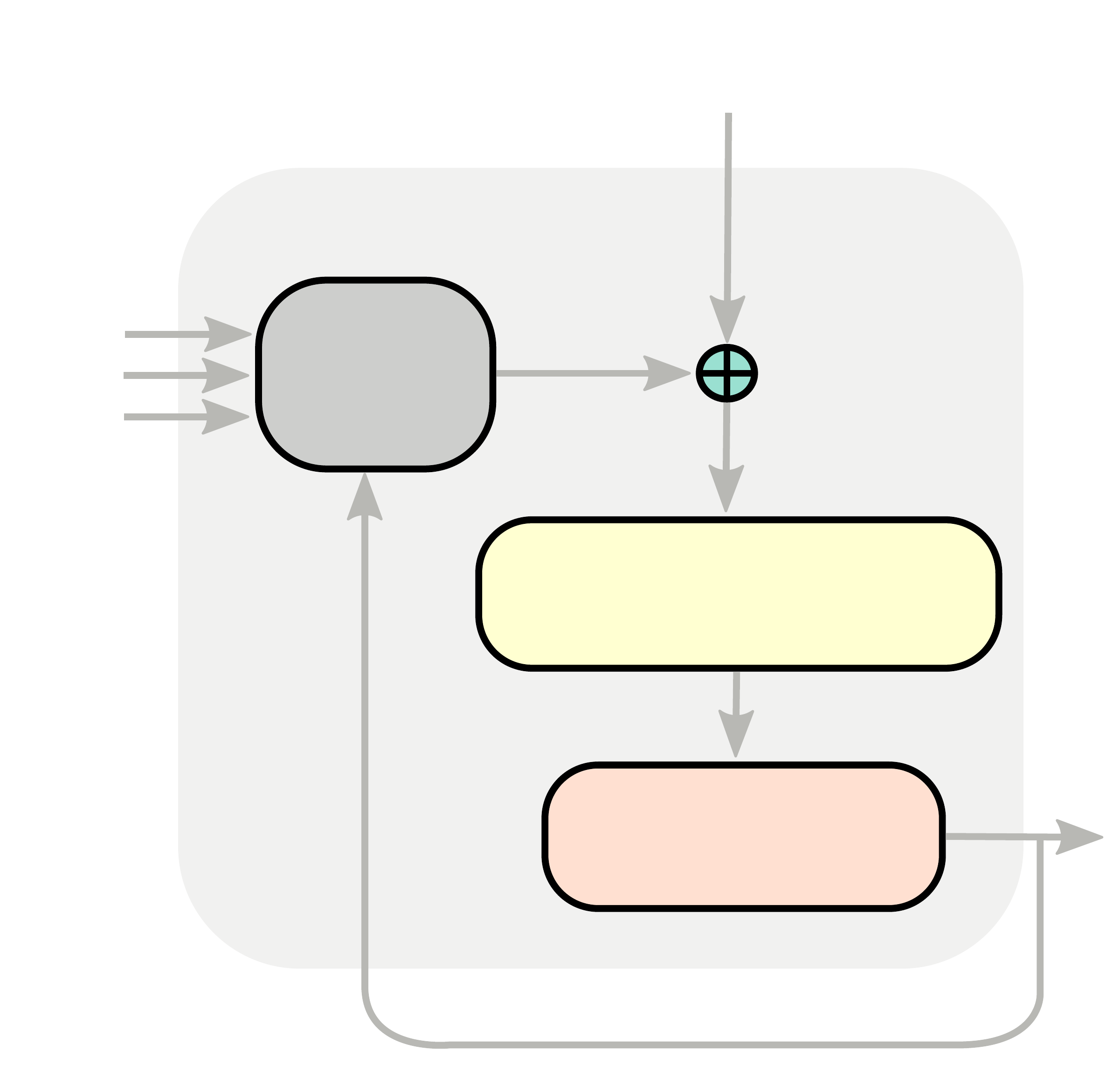}}; 
        \end{scope}
        \node (note) at (-1,1.1) {NN};
        \node (note) at (-0.95,0.75) {(<1 MB)};
        \node (note) at (1,-0.25) {Cost, $\frac{\partial Cost}{\partial \mathbf{u}}$, $\frac{\partial^2 Cost}{\partial \mathbf{u}^2}$};
        \node (note) at (1,-1.52) {NR Solver};
        \node (note) at (2.8,-2) {$\mathbf{u}^*$};
        \node (note) at (-2.7,1.2) {$\hat{\mathbf{y}}_{\text{queue}}$};
        \node (note) at (-2.7,0.90) {$\mathbf{u^*}_{\text{queue}}$};
        \node (note) at (-2.7,0.60) {$\mathbf{l}_t$};
        \node (note) at (1.3,2.2) {$\mathbf{y}_{ref}$};
        \node (note) at (0.2,1.1) {$\hat{\mathbf{y}}$};
         \node (note) at (-2,2.1) {\textit{\textbf{(B)}}};
        \end{tikzpicture}
    \end{subfigure}
    \hspace{-20pt}
    \begin{subfigure}[t]{0.5\columnwidth}
        \centering
        \begin{tikzpicture}
        \begin{scope}
        \node (image) at (0, 0) {\includegraphics[height=2.2 in]{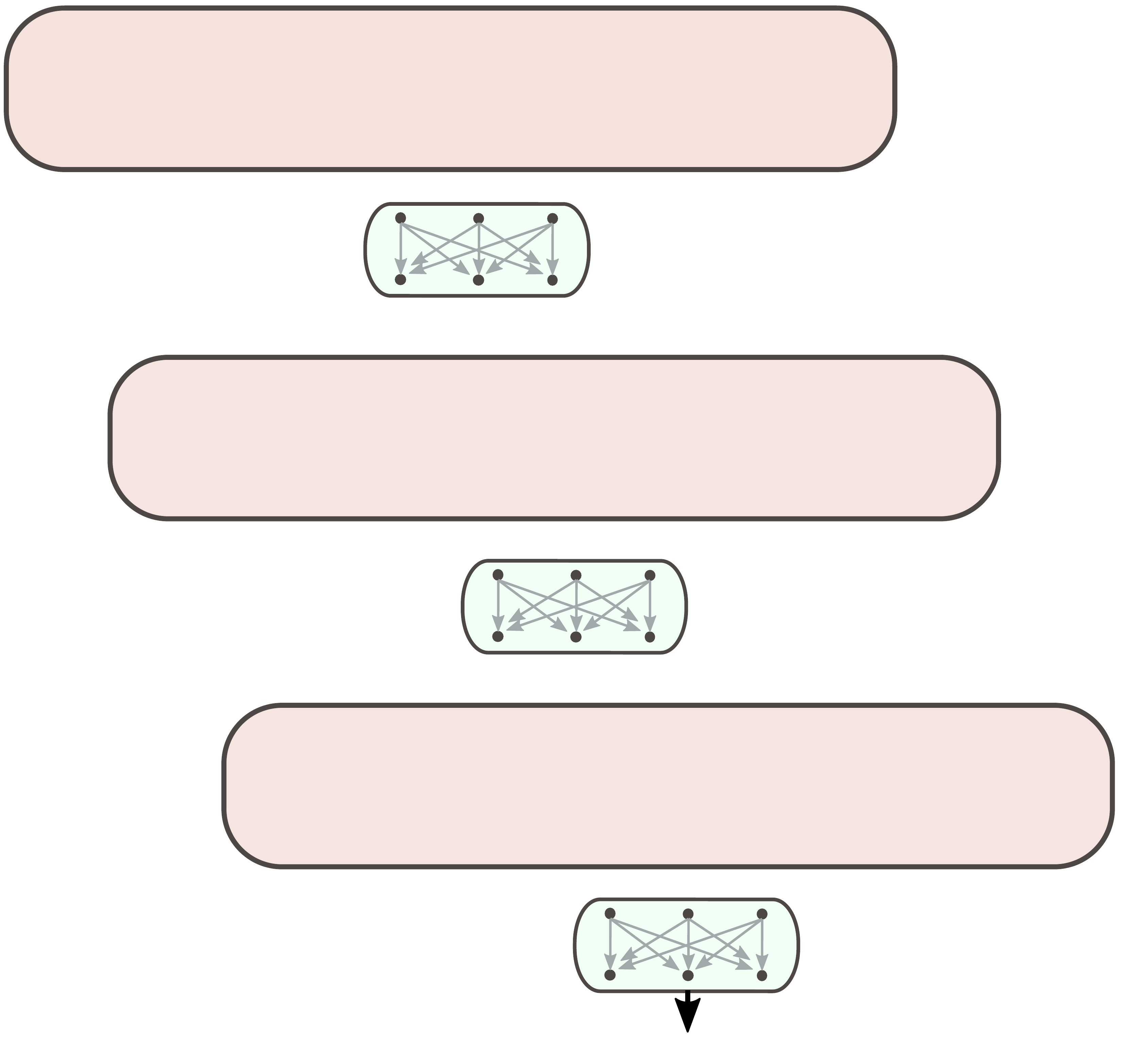}};
        \end{scope}
        \node (note) at (-3.3,2.1) {\textit{\textbf{(C)}}};
         
         \node (note) at (-1.9,2.3) {$\mathbf{u}_{t}, ... ,\mathbf{u}_{t-n_d},$};
         \node (note) at (0.1,2.3) {$\mathbf{y}_{t-1}, ... ,\mathbf{y}_{t-d_d},$};
         \node (note) at (1.3,2.3) {$\mathbf{l}_t$};
         
         \node (note) at (-1.3,.45) {$\mathbf{u}_{t+1}, ... ,\mathbf{u}_{t-n_d+1},$};
         \node (note) at (0.8,.45) {$\highlight{\mathbf{y}_{t}}, ... ,\mathbf{y}_{t-d_d+1},$};
         \node (note) at (1.9,.45) {$\mathbf{l}_t$};
         
         \node (note) at (-0.25,-1.2) {$\mathbf{u}_{t+N_c-1}, ... ,\mathbf{u}_{t-n_d+N_c-1},$};
         \node (note) at (0.65,-1.6) {$\mathbf{y}_{t +N -d_d}, ... ,\highlight{\mathbf{y}_{t+N-1}},$};
         \node (note) at (2.15,-1.6) {$\mathbf{l}_t$};
         
         \node (note) at (0.6,-3.0) {$\highlight{\mathbf{y}_{t+N}}$};
          \node (note) at (0.3,1.5) {$h$};
          \node (note) at (0.8,-0.4) {$h$};
          \node (note) at (1.4,-2.3) {$h$};
        \end{tikzpicture}
    \end{subfigure}
    \caption{\textit{(A)} The steps from data collection to offloading the controller.  \textit{(B)} Controller description. (\textit{C}) Recursive neural network predictions.}
    \label{fig:pipeline}
    \vspace{-10pt}
\end{figure}

\begin{figure}[H]
    \hfill
     \begin{subfigure}[t]{\columnwidth}
        \centering
        \begin{tikzpicture}
        \begin{scope}
        \node (image) at (0,0) {\includegraphics[width=1
        \columnwidth, trim={0cm 0cm 0cm 0cm},clip]{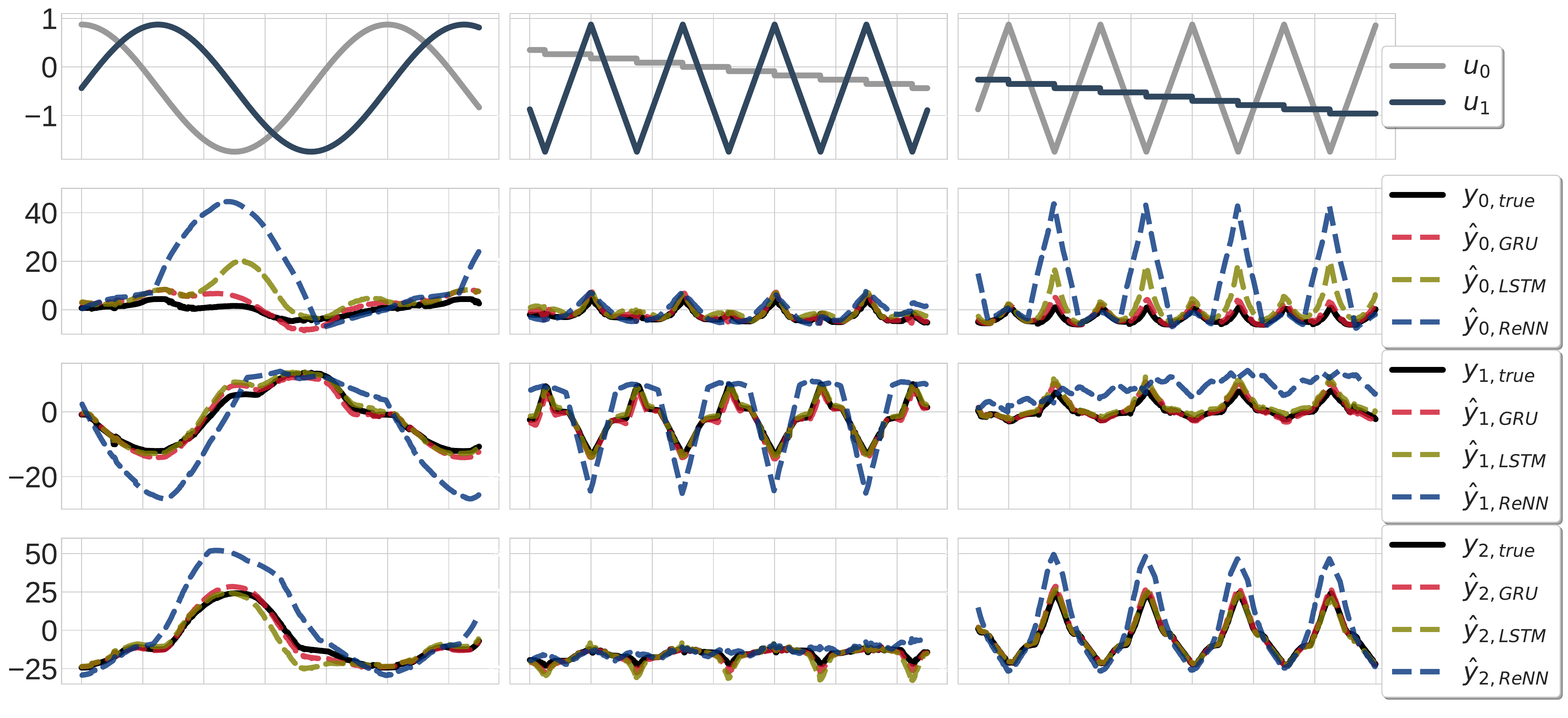}};
        \end{scope}
        \node (note) at (-.5,2.8) {\textbf{Testing Set Predictions at Different Stages of the Data Collection Sequence}};
        \node[rotate=90] (note) at (-6,0.7) {$y_0$ [mm]};
        \node[rotate=90] (note) at (-6,-0.65) {$y_1$ [mm]};
        \node[rotate=90] (note) at (-6,-2.2) {$y_2$ [mm]};
        \node[rotate=90] (note) at (-6,2.25) {$u$ [rad]};
        \end{tikzpicture}
    \end{subfigure}
    \hfill
      \vspace{-20pt}
    \caption{Sample of the sequence swept by the block for data collection.}
    \label{fig:model}
    \vspace{-10pt}
\end{figure}
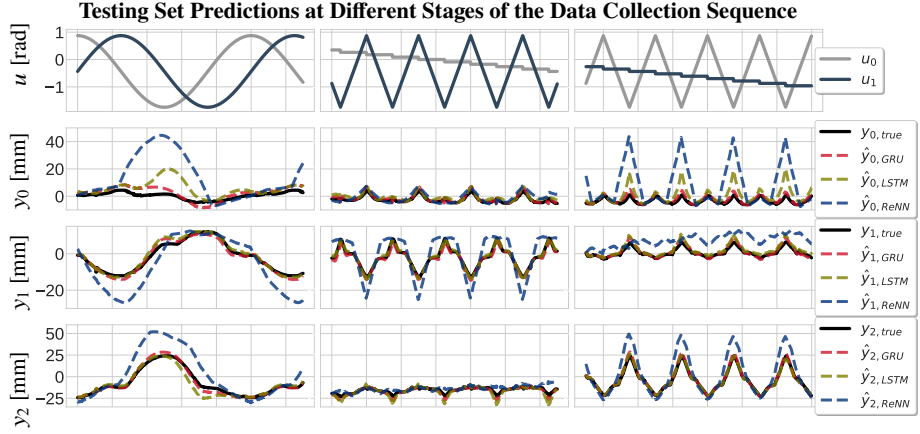

 In this work, we compare three different types of recursive neural network (RecNN) models that were each trained using 10-fold time series split. These models are recursive because the output of the model feeds back in part of the input of the next time step in a tree-like structure that reuses the weights as seen in Fig. \ref{fig:pipeline}\textit{(C)}. This requires that we define a trainable child structure for the neural network, which we call $h$. We test three model representations of this neural network sub-structure: A fully connected substructure \textit{h= [Dense(5, relu), Dense(3, tanh)]}, that is similar to an RNN where the recurrent weights are the same as the feedforward ones; an LSTM-based child model trained is given by \textit{h= [LSTM(5), Dense(5, tanh), Dense(3, tanh)]} and the GRU  trained is given by \textit{h= [GRU(5), Dense(5, relu), Dense(3, linear)]}. These lead to $243$, $435$, and $570$ total model parameters respectively, and get recursively called $N$ times each prediction step.
 These model sizes require flash memory as low as $0.97kB$, $1.7kB$ and $2.22kB$ using $float32$. A sample of the off-line testing set results are described in Fig. \ref{fig:model}, and the $L_2$ errors over the 10 partitions are $2.52 \pm 1.52mm$ for the recurrent model; $0.26\pm 0.04 mm$ for the GRU model and $0.26 \pm 0.04mm$ for the LSTM model. 

\runinhead{Cost Function}

The cost function $J$ used in this work is a combination between the standard, unconstrained, MPC cost function with additional terms that enforce smoothness in the optimal control inputs found in \cite{ngpc}. It is displayed in Eq. \ref{eq:cost}. $\mathbf{\Lambda} \in \mathbb{S}^{m\times m}$  and $\mathbf{Q} \in \mathbb{S}^{n \times n}$ are weighting matrices for the input changes and output square errors, respectively. The subscript $ref$ denotes the target path to follow. 
\vspace{-3pt}
\begin{equation}
    \label{eq:cost}
    \begin{aligned}
        J & = \sum_{j=N_1}^{N_2} \| \mathbf{y}_{ref, j} - \hat{\mathbf{y}}_j \|_{\mathbf{Q}}^2   + \sum_{j=0}^{N_c} \| \Delta \mathbf{u}_j \|_{\mathbf{\Lambda}}^{2} \\ & + \sum_{i=1}^{m}  \sum_{j=1}^{N_c} \Bigg[ \frac{s}{u(n+j, i) + \frac{r}{2} - b} + \frac{s}{\frac{r}{2} + b - u(n+j, i)} - \frac{4}{r} \Bigg] 
    \end{aligned}
\end{equation}

\runinhead{Newton-Raphson Solver} In order to get an optimal control input, we frame the control problem as the solution to Eq. \ref{eq:NR}, where $\frac{\partial^2 J}{\partial \mathbf{U}^2}(k)$ is the Hessian of the cost function at the current timestep $k$, and $\frac{\partial J}{\partial \mathbf{U}}(k)$ is the Jacobian of the cost function. This approach was first formulated in \cite{ngpc}, where $\mathbf{U}$ as the vector composed by $\mathbf{U} = \{ \mathbf{u}_k, \mathbf{u}_{k+1}, ..., \mathbf{u}_{k+N_c}\}^T \in \mathbb{R}^{N_c \times m}$.

\begin{equation}
    \label{eq:NR}
    \begin{aligned}
    \frac{\partial^2 J}{\partial \mathbf{U}^2}(k)(\mathbf{U}(t+1) - \mathbf{U}(k)) = -  \frac{\partial J}{\partial \mathbf{U}}(k)
    \end{aligned}
\end{equation}

    \vspace{-15pt}
\begin{equation}
    \label{eq:cost1}
    \begin{aligned}
        \frac{\partial J}{\partial \mathbf{U}}(k) & \approx -2 \sum_{j=N_1}^{N_2} (\mathbf{y}_{ref, j} - \hat{\mathbf{y}}_j)^T \mathbf{Q} \overbrace{\frac{\partial \hat{\mathbf{y}}_j}{\partial \mathbf{U}}}^{\text{Eqn. \ref{eq:first_derivative}}} + 2 \sum_{j=0}^{N_c} \Delta \mathbf{u}_j \mathbf{\Lambda} \frac{\partial \Delta \mathbf{u}_j}{\partial \mathbf{U}} \\ & + \sum_{i=1}^{m}  \sum_{j=1}^{N_c}  \Bigg[ \frac{-s}{\Big[ u(n+j, i) + \frac{r}{2} - b  \Big]^2} + \frac{s}{\Big[ \frac{r}{2} + b - u(n+j, i) \Big]^2} \Bigg] \in \mathbb{R}^{N_c \times m}
    \end{aligned}
\end{equation}
\vspace{-8pt}
\begin{equation}
    \label{eq:cost2}
    \begin{aligned}
      \frac{\partial^2 J}{\partial \mathbf{U}^2}(k) & \approx 2 \sum_{j=N_1}^{N_2}   \Big[ \mathbf{Q}  \Big(\frac{\partial \hat{\mathbf{y}}_j}{\partial \mathbf{U}} \circ \frac{\partial \hat{\mathbf{y}}_j}{\partial \mathbf{U}} \Big) -  (\mathbf{y}_{ref, j} - \hat{\mathbf{y}}_j)^T \mathbf{Q} \frac{\partial ^2 \hat{\mathbf{y}}}{\partial \mathbf{U}^2} \Big] + \\ & 2\sum_{j=0}^{N_c} \Big[ \mathbf{\Lambda}  \Big( \frac{\partial \Delta \mathbf{u}_j}{\partial \mathbf{U}} \circ \frac{\partial \Delta \mathbf{u}_j}{\partial \mathbf{U}} \Big)  + \Delta \mathbf{u}_j \mathbf{\Lambda} \frac{\partial ^2 \Delta \mathbf{u}_j}{\partial \mathbf{U}^2} \Big] \\ & 
        + \sum_{i=1}^{m}  \sum_{j=1}^{N_c}  \Bigg[ \frac{2s}{\Big[ u(n+j, i) + \frac{r}{2} - b  \Big]^3} + \frac{2s}{\Big[ \frac{r}{2} + b - u(n+j, i) \Big]^3} \Bigg] \in \mathbb{R}^{N_c \times N_c}
    \end{aligned}
\end{equation}

Eq. \ref{eq:cost1} and \ref{eq:cost2} describe the expressions used toward the Jacobian and Hessian of the cost in Eq. \ref{eq:cost}. Let $p = n_dm + d_dn+w$ be the length of the flattened input of the neural network and $\varepsilon$ be the differentiation stencil step length such that $[\mathbf{x}_{inputs} + \varepsilon\mathbf{I}] \in \mathbb{R}^{p \times p}$.
\vspace{-5pt}
\begin{equation}
    \label{eq:first_derivative1}
   \Theta =  \frac{\partial \hat{\mathbf{y}}}{\partial \mathbf{x}_{inputs}} =   \frac{g(\mathbf{x}_{inputs} + \varepsilon\mathbf{I})) - g(\mathbf{x}_{inputs} - \varepsilon\mathbf{I})}{2\varepsilon} + \mathcal{O}(\varepsilon^2) \in  \mathbb{R}^{p \times n}
\end{equation}

\begin{equation}
    \label{eq:first_derivative}
    \frac{\partial \hat{\mathbf{y}}}{\partial \mathbf{U}} \approx \begin{bmatrix}  \sum_k^{n_d} \Theta_{mk+0, :} , & ... & ,\sum_k^{n_d} \Theta_{mk+(m-1), :} 
    \end{bmatrix}^T \in \mathbb{R}^{n \times m}
\end{equation}

For the second derivative ($\frac{\partial ^2 \hat{\mathbf{y}}}{\partial \mathbf{U}^2}$) we do a similar treatment to obtain this matrix, the resulting matrix is in $\mathbb{R}^{m \times m}$ using a second order finite difference stencil for the second derivative. This requires that the first elements of the input vector of the neural network correspond to the control inputs as seen in Fig. \ref{fig:pipeline}\textit{(C)}.

\vspace{-10pt}

\section{Results}
\label{sec:results}

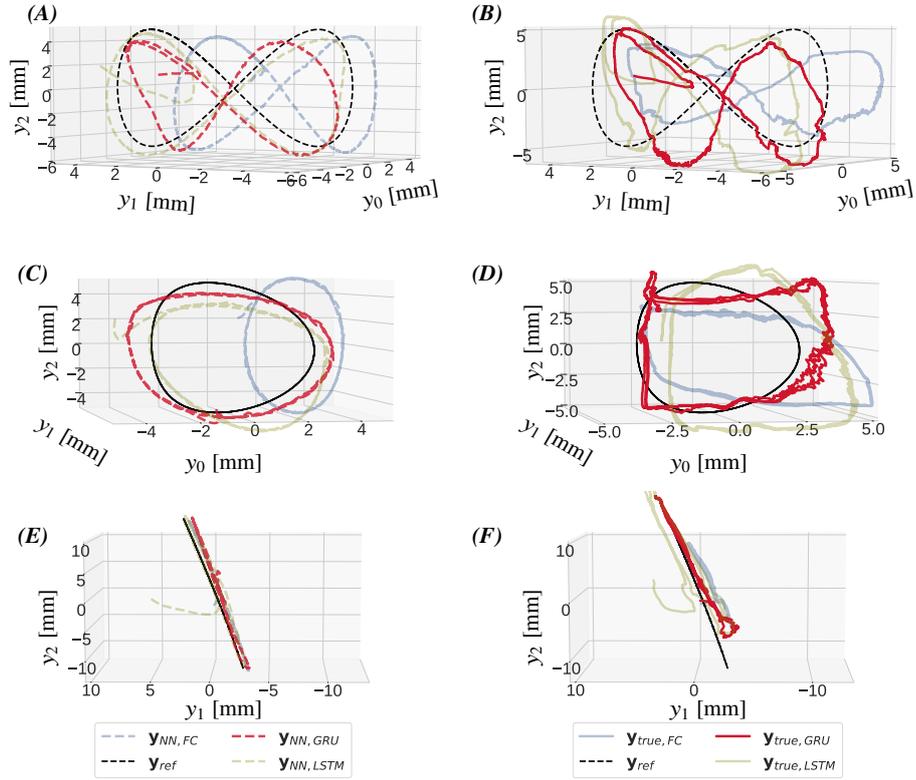
\begin{figure}[H]
    \vspace{-5pt}
    \centering
    \begin{subfigure}[t]{0.48\columnwidth}
        \centering
        \begin{tikzpicture}
        \begin{scope}
        \node (image) at (0,0) {\includegraphics[width=\columnwidth, trim={6cm 1cm 6cm 1cm},clip]{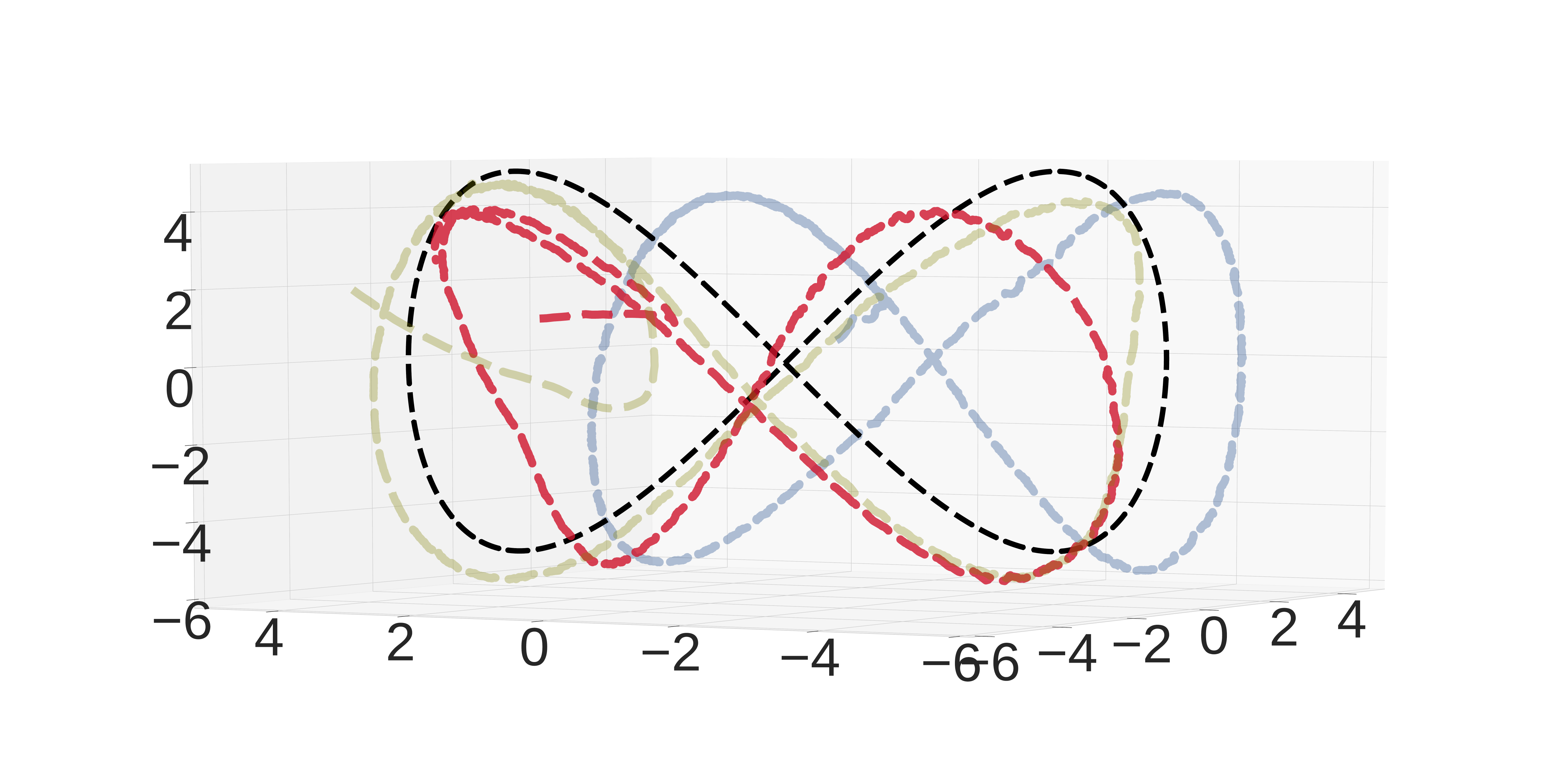}}; 
        \end{scope}
        \node[rotate=90] (note) at (-2.8,0.0) {$y_2$ [mm]};
        \node[rotate=-5] (note) at (-1.,-1.4) {$y_1$ [mm]};
         \node[rotate=10] (note) at (2.2,-1.3) {$y_0$ [mm]};
         \node (note) at (-2.5,1.1) {\textit{\textbf{(A)}}};
        \end{tikzpicture}
    \end{subfigure}
    \hfill
    \begin{subfigure}[t]{0.48\columnwidth}
        \centering
        \begin{tikzpicture}
        \begin{scope}
        \node (image) at (0,0) {\includegraphics[width=\columnwidth, trim={6cm 1cm 6cm 1cm},clip]{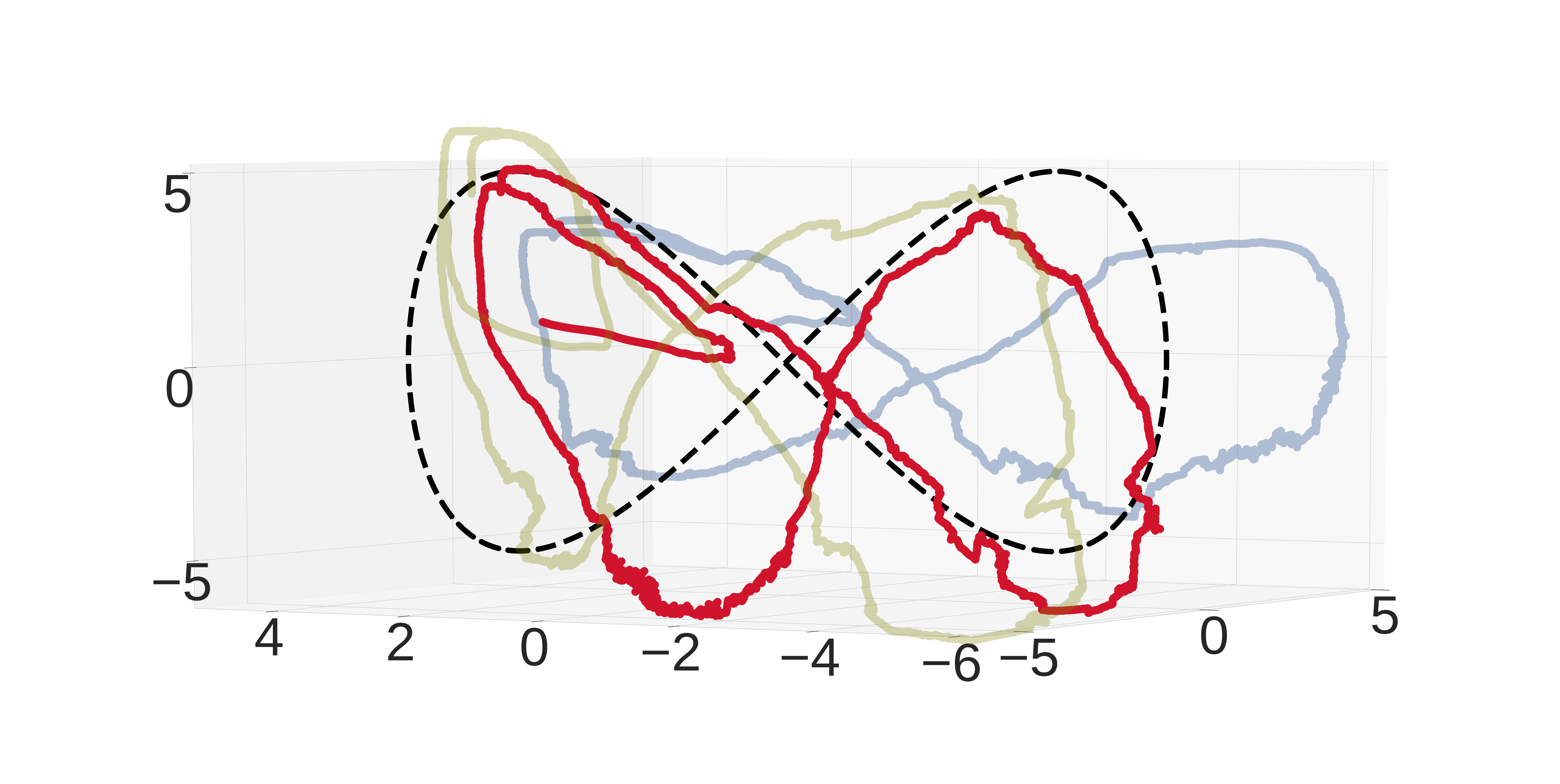}}; 
        \end{scope}
       \node[rotate=90] (note) at (-2.8,0.) {$y_2$ [mm]};
        \node[rotate=-5] (note) at (-1,-1.4) {$y_1$ [mm]};
         \node[rotate=10] (note) at (2.2,-1.3) {$y_0$ [mm]};
          \node (note) at (-2.9,1.1) {\textit{\textbf{(B)}}};
        \end{tikzpicture}
    \end{subfigure}
    \centering
    \begin{subfigure}[t]{0.48\columnwidth}
        \centering
        \begin{tikzpicture}
        \begin{scope}
        \node (image) at (0,0) {\includegraphics[width=\columnwidth, trim={6cm 1cm 6cm 1cm},clip]{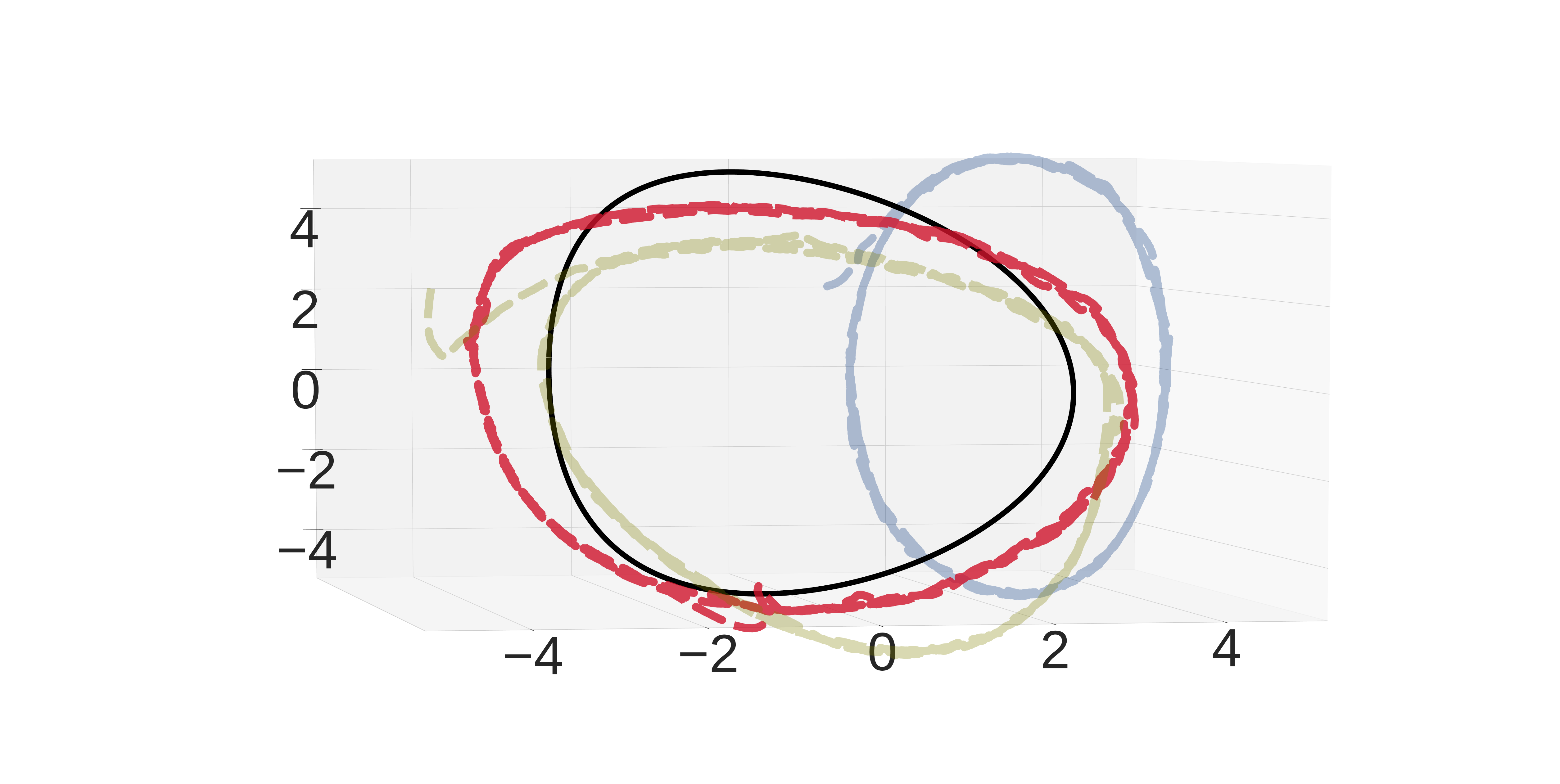}}; 
        \end{scope}
       \node[rotate=90] (note) at (-2.3,0.1) {$y_2$ [mm]};
        \node[rotate=0] (note) at (0.0,-1.5) {$y_0$ [mm]};
         \node[rotate=-30] (note) at (-2.,-1.2) {$y_1$ [mm]};
         \node (note) at (-2.5,1.0) {\textit{\textbf{(C)}}};
        \end{tikzpicture}
    \end{subfigure}
    \hfill
    \begin{subfigure}[t]{0.48\columnwidth}
        \centering
        \begin{tikzpicture}
        \begin{scope}
        \node (image) at (0,0) {\includegraphics[width=\columnwidth, trim={6cm 1cm 6cm 1cm},clip]{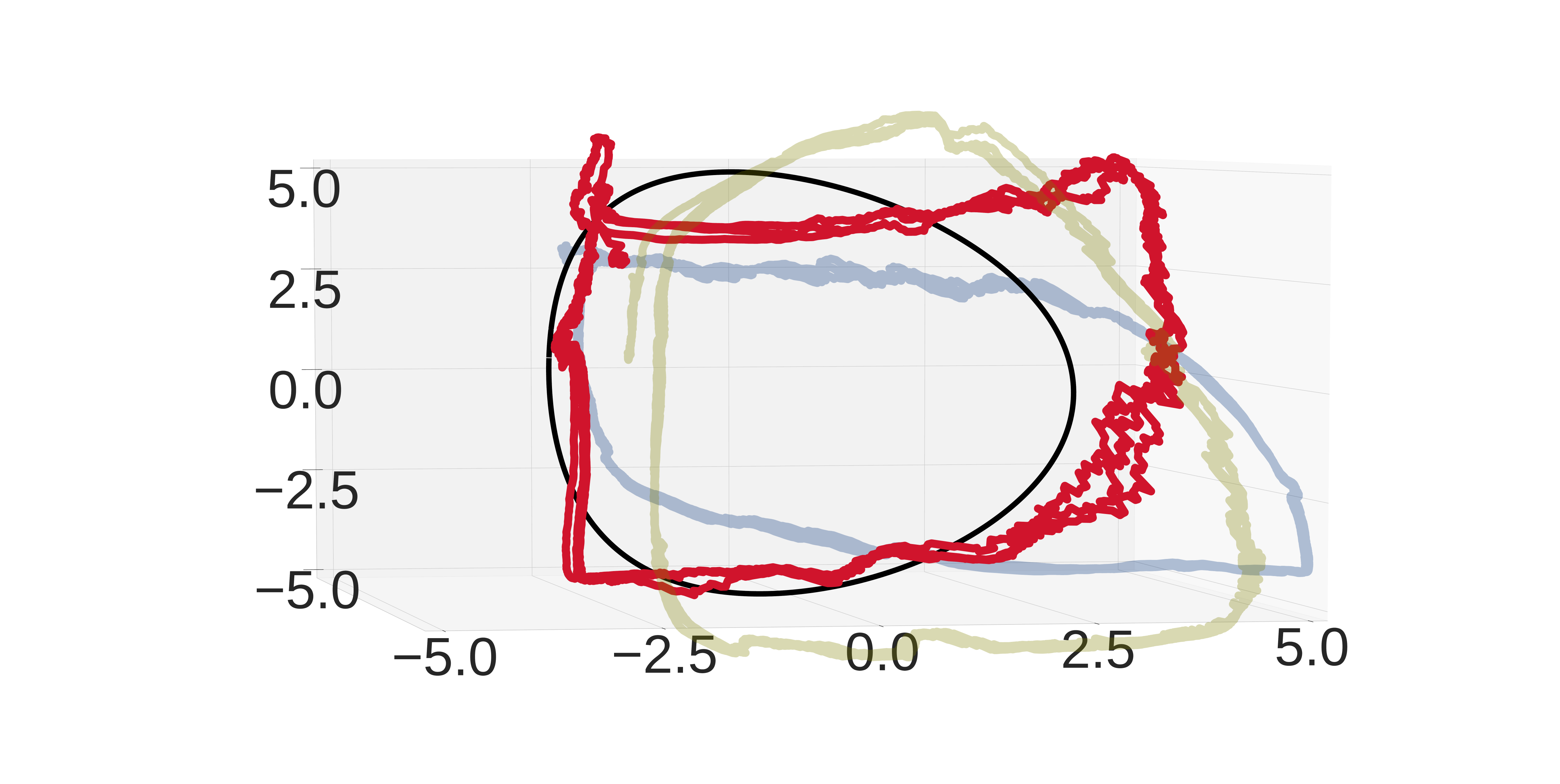}}; 
        \end{scope}
        \node[rotate=90] (note) at (-2.3,0.1) {$y_2$ [mm]};
        \node[rotate=0] (note) at (0.0,-1.5) {$y_0$ [mm]};
         \node[rotate=-30] (note) at (-2.,-1.2) {$y_1$ [mm]};
          \node (note) at (-2.9,1.0) {\textit{\textbf{(D)}}};
        \end{tikzpicture}
    \end{subfigure}
    \centering
        \begin{subfigure}[t]{0.48\columnwidth}
        \centering
        \begin{tikzpicture}
        \begin{scope}
        \node (image) at (0,0) {\includegraphics[width=\columnwidth, trim={6cm 0cm 6cm 1cm},clip]{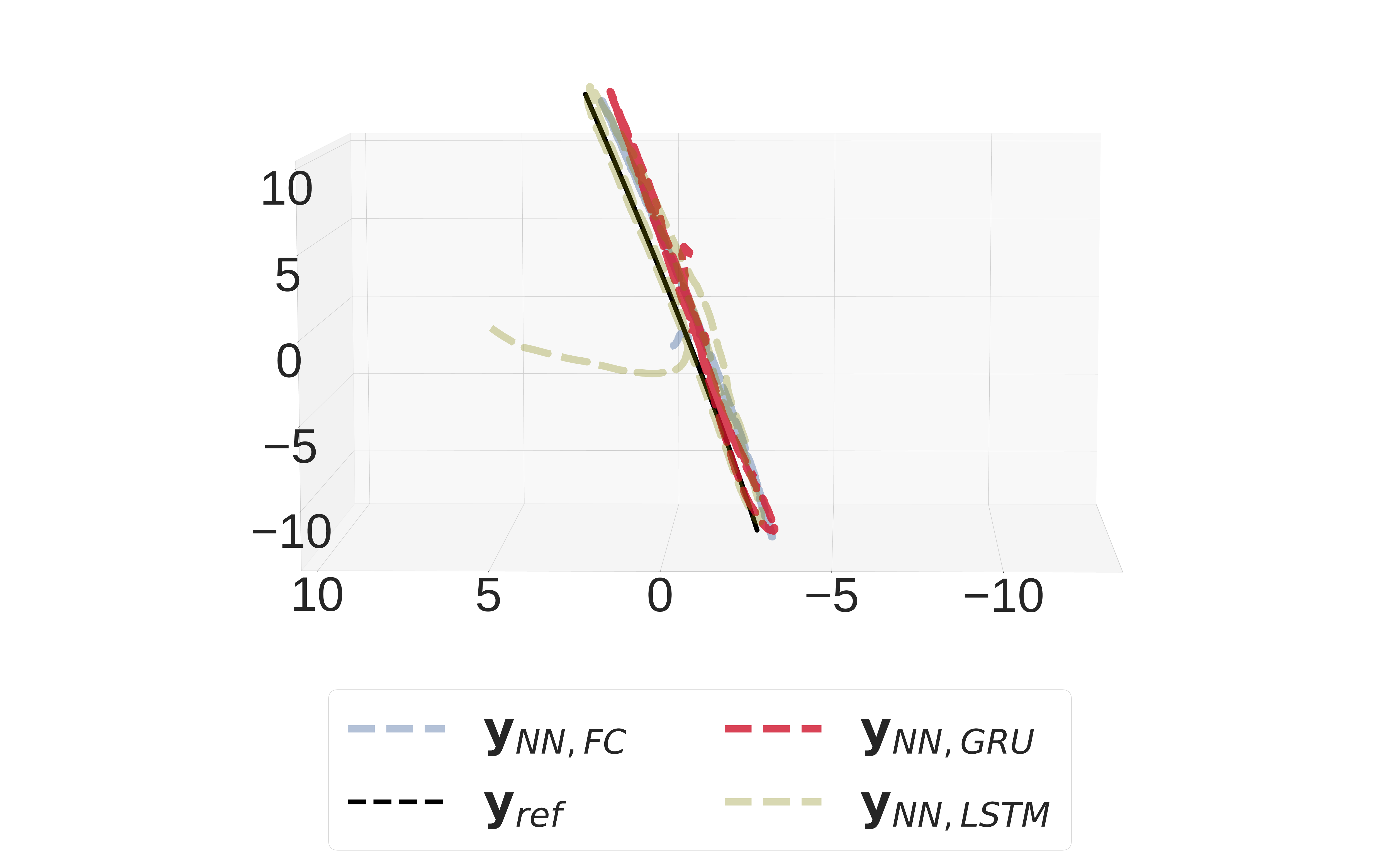}}; 
        \end{scope}
        \node[rotate=90] (note) at (-2.3,0.1) {$y_2$ [mm]};
        \node[rotate=0] (note) at (0.0,-1.) {$y_1$ [mm]};
         \node (note) at (-2.5,1.3) {\textit{\textbf{(E)}}};
        \end{tikzpicture}
    \end{subfigure}
    \hfill
    \begin{subfigure}[t]{0.48\columnwidth}
        \centering
        \begin{tikzpicture}
        \begin{scope}
        \node (image) at (0,0) {\includegraphics[width=\columnwidth, trim={6cm 0cm 6cm 1cm},clip]{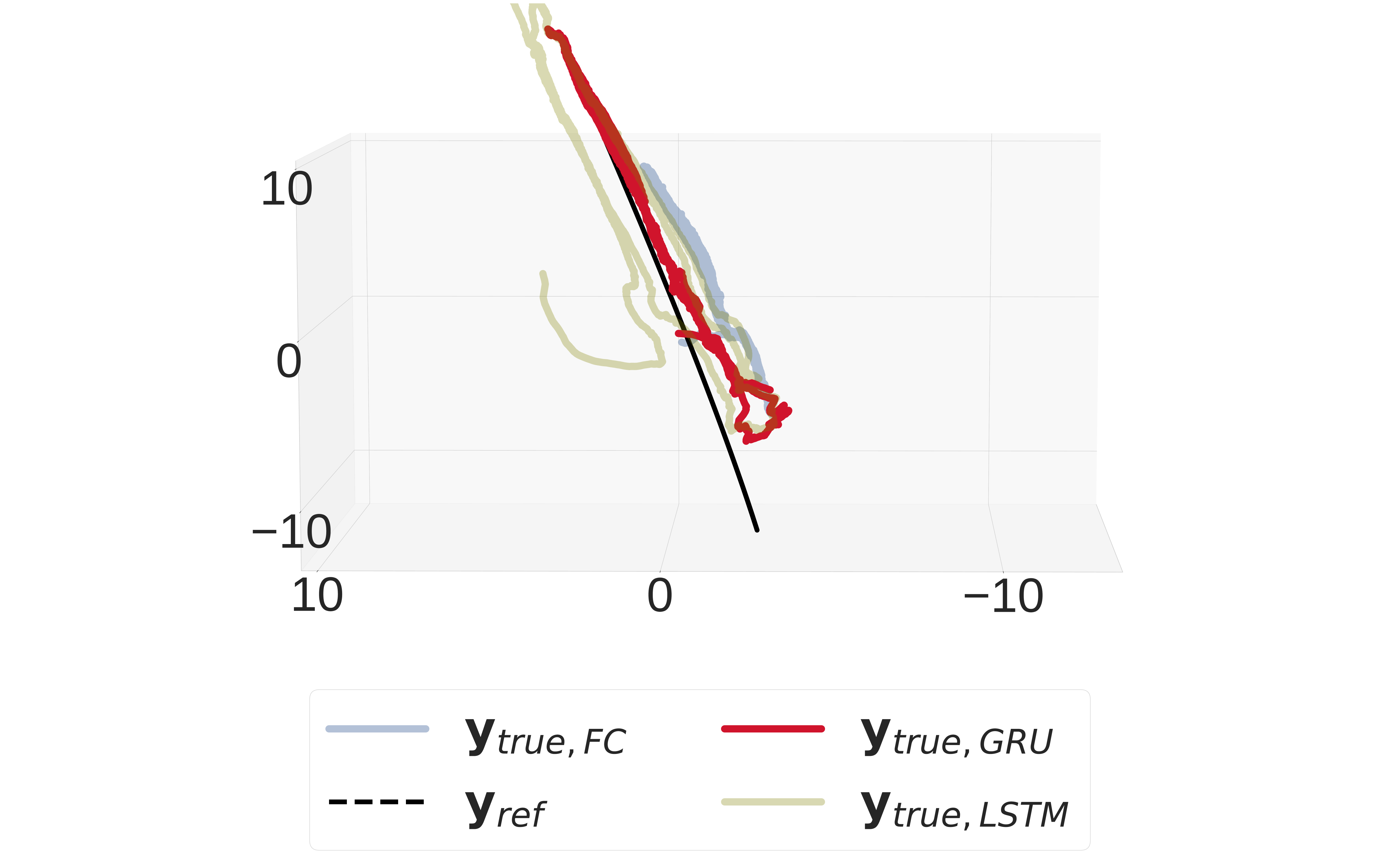}}; 
        \end{scope}
       \node[rotate=90] (note) at (-2.3,0.1) {$y_2$ [mm]};
        \node[rotate=0] (note) at (0.0,-1.) {$y_1$ [mm]};
          \node (note) at (-2.9,1.3) {\textit{\textbf{(F)}}};
        \end{tikzpicture}
    \end{subfigure}
    \caption{3D trajectory tracking. Left column (A, C, E) indicates the mean predicted trajectories and the right column (B, D, F) indicates the mean of the true trajectories. Best performing model from Table \ref{tab:1} is highlighted in each task.}
    \label{fig:path_tracking}
    \vspace{-10pt}
\end{figure}

We test our controller on the path following tasks that we named: 1) \textit{Eight}, 2) \textit{Pringle} and 3) \textit{Line}, and which are defined below. We describe the \textit{Eight} reference trajectory as $\mathbf{y_{ref}}(t) = 
\{A sin^2(\omega t ) + y_{0, 0}, 
 B sin(\omega t ) cos(\omega t ) + y_{1, 0},  
 C sin(\omega t ) + y_{2, 0} \}^T$
 , where $\omega$ 
 is the frequency of the wave described by this path. Parameters $A$ and $B$ are unitless multipliers that indicate amplitude and extension of the reference path geometry. Fig. \ref{fig:path_tracking} shows the difference between the true location of the end effector compared with the estimated location of the end effector for three different recurrent models used in this work. Tab. \ref{tab:1} summarizes the root mean squared error statistics over 10 runs of each of the paths described. \textit{Pringle} is defined as a hyperbolic paraboloid $\mathbf{y_{ref}}(t) = \{
A (y_2^2 / B^2 - y_1^2 / C^2) + y_{0, 0},
Bcos(2\pi  \omega) t + y_{1, 0}, 
Csin(2\pi  \omega) t + y_{2, 0}\}^T$, 
and \textit{Line} is defined as $\mathbf{y_{ref}}(t) = 
\{A (y_2^2 / B^2 - y_1^2 / C^2) + y_{0, 0}, 
Bsin(2\pi  \omega t+ 10^{-6} t^2) + y_{1, 0}, 
Csin(2\pi  \omega t+ 10^{-6} t^2) + y_{2, 0} \}^T$.

\begin{table}[!htbp]
\caption{Summary of the RMSE statistics in three different paths.}
\label{tab:1}    
\vspace{-7pt}
\begin{tabular}{p{1cm}p{2cm}p{2.5cm}p{2.5cm}p{2.5cm}}
\hline\noalign{\smallskip}
Model & $n^{\circ}$ Parameters$^a$  & \textit{Eight} & \textit{Pringle} & \textit{Line} \\
\noalign{\smallskip}\svhline\noalign{\smallskip}
FC  & $243$ & $1.36 \pm 0.21 mm$  & $1.93 \pm 0.48 mm$ &
$2.02 \pm 1.24 mm$\\
GRU   & $435$ & $1.29 \pm 0.28 mm$ & $1.56 \pm 0.58 mm$ & $2.01\pm 1.19 mm$\\
LSTM  & $570 $ & $1.42 \pm 0.39 mm$  & $2.35 \pm 0.23 mm$ & $2.56 \pm 1.44 mm$\\
\noalign{\smallskip}\hline\noalign{\smallskip}
\end{tabular} \\
$^a$ total number of parameters in the neural network, counting weights and biases.
\vspace{-10pt}
\end{table}

\vspace{-10pt}

\runinhead{Microcontroller Considerations} We timed the computation required to perform model prediction, first and second derivatives of the model in an Tensilica Xtensa LX6 microprocessor (ESP32, Espressif Systems) in the Arduino IDE for $N=3$ and obtained around $0.25 ms$, $5.86 ms$ for Eq. \ref{eq:first_derivative1} and $9.07 ms$ for prediction, first derivative and second derivative, averaged over $5,000$ steps using \textit{nn4mc} \cite{nn4mc} on the FC model presented in this work. When we use shared operations for the two derivatives we obtain an average time of $11.35 ms$ to obtain the two derivatives. These preliminary results are promising to implement the algorithm presented in commodity microcontrollers.

\begin{figure}[H]
    \hspace{-28pt}
    \begin{tikzpicture}
        \begin{scope}[shift={(0,0)}]
    \node (image) at (0,0) {\includegraphics[width = \columnwidth, trim={0cm 0cm 0cm 0cm},clip]{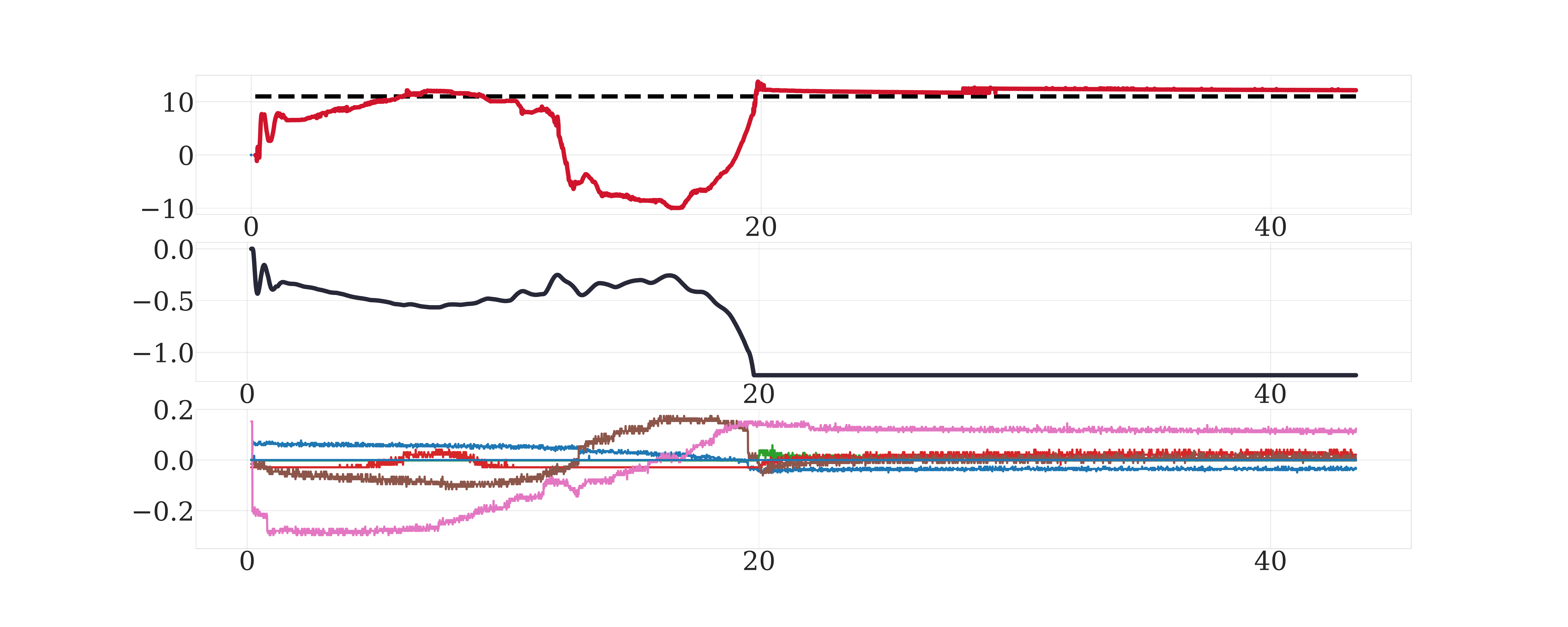}};
     \node[rotate=0] (note) at (0.0,-2.1) {Time [s]};
  \node[rotate=90] (note) at (-5.,-1.1) {$l - \bar{l}$};
   \node[rotate=90] (note) at (-5.,-0.0) {$u_0 [rad]$};
    \node[rotate=90] (note) at (-5.,1.4) {$y_2 [mm]$};
    \node[rotate=0] (note) at (0,2.0) {\textbf{Response of the Controller to Mechanical and EM disturbance}};
    \draw [dashed] (-2.1,-1.8) -- (-2.1,1.8);
    \node[rotate=0] (note) at (-2.,-2.) {$+137g$};
    
    \node (image) at (5.5,1.8) {\includegraphics[width = 0.8 in, trim={0cm 0cm 0cm 0cm},clip]{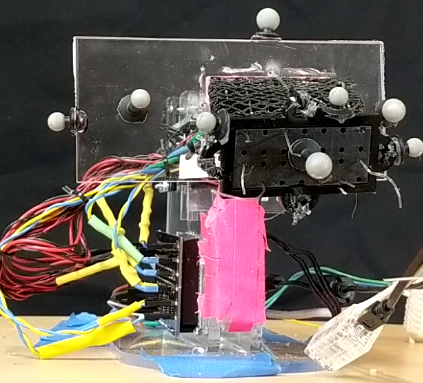}};
 \node (image) at (5.5,0.0) {\includegraphics[width= 0.8 in, trim={0cm 0cm 0cm 0cm},clip]{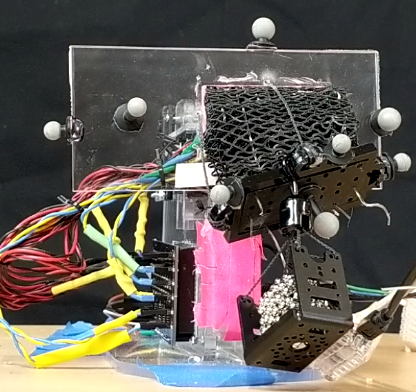}};
 \node (image) at (5.5,-1.8) {\includegraphics[width = 0.8 in, trim={0cm 0cm 0cm 0cm},clip]{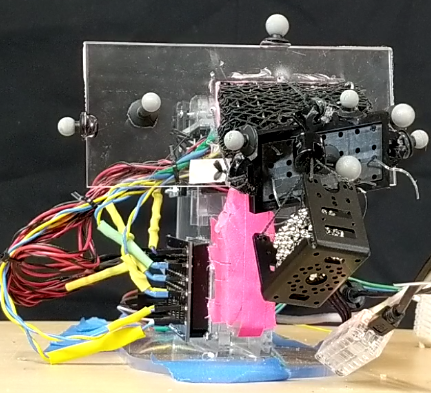}};
    \end{scope}
    \end{tikzpicture}
    \vspace{-10pt}
    \caption{Disturbance response. Top figure indicates the true position in millimeters. Second figure indicates the changes in servo command that the controller solved for. Bottom figure illustrates changes in light lace sensor signals.}
    \label{fig:disturbance}
    \vspace{-10pt}
\end{figure}
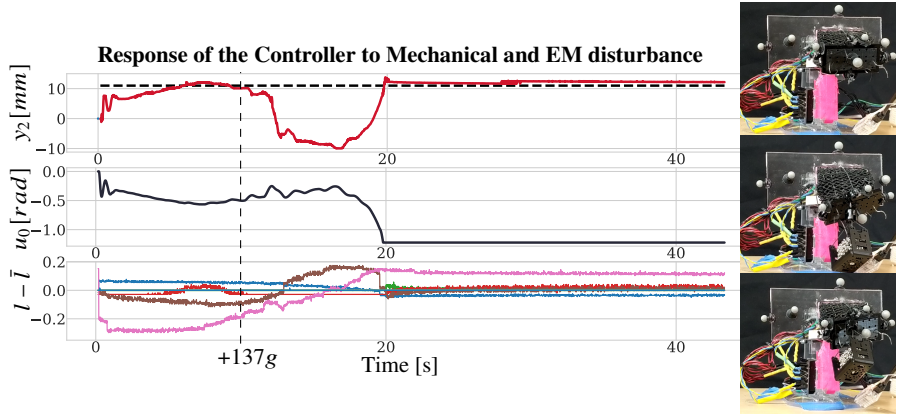

\runinhead{Mechanical and Electromagnetic Disturbance Rejection.} We expose the end effector to mechanical and electromagnetic disturbances by loading magnetic balls (Nickel Set, Speks) with a total mass of $100g$ at the end effector position using a $37g$ crane to hold them. Fig. \ref{fig:disturbance} displays the response of both the sensors and the controller to the presence of such disturbances. 

\runinhead{Discussion}

From the results displayed in Sec. \ref{sec:results}, despite the inaccuracies in the model prediction found in Fig. \ref{fig:model}, the optimization algorithm solves for optimal control inputs using the predicted position coming from the neural network. This means that the control performance is associated with the model prediction performance, thus the better prediction power the model has, the higher the control accuracy will be. In Fig \ref{fig:path_tracking} (C) we observe that given the increased model prediction errors in the FC model in the $y_0$ direction, the prediction curve projects onto another plane, but we obtain a motion close to the desired motion by assigning a lower diagonal value ($1e-3$) in the first row of $\mathbf{Q}$ for the FC controller, which decreases the penalization of errors in that direction. The GRU-based model predicted the off-line testing set data closer and this reflects on the accuracy of the controller based on this model, where we evidence an increased performance relative to the other models compared. We find that even though the past history of inputs and outputs is important for the computation of the control input, when we confound this with the true measurements during the disturbance rejection experiment, models and controller proved to respond to changes in the signal coming from the sensors. This is evidenced by the changes in optimal control input signals during the disturbance rejection experiment in Fig. \ref{fig:disturbance}.

\runinhead{Limitations to this Approach} This approach is most effective when the  end effector can physically reach the reference path. To prevent the servos from saturation, we clip the optimal control input, servo inputs did not reach saturation during the experiments in Fig. \ref{fig:path_tracking}, but during the tuning process.  This algorithm is most effective with knowledge of the initial state of the end effector for better initial neural network predictions. This initialization can be consistent throughout the experiments. Mechanical disturbance rejection is most effectively achieved when the sensor normalization allows for very sensitive sensor reading patterns. Therefore, the calibration used in Fig. \ref{fig:disturbance} was different than that of the experiments in Fig. \ref{fig:path_tracking}.

\runinhead{Conclusion}

We use parsimonious neural network models in a hierarchical, recursive configuration to predict the forward kinematics of the end effector and develop a nonlinear predictive controller to optimize the control inputs. We achieve sub-centimeter accuracy that compares favorably with those in the related literature. We conclude that light lace networks \cite{optical} are a reliable sensing modality that is not affected by electromagnetic and moderate mechanical disturbances and can be used for feedback control in truly soft actuation systems. Future work includes embedding the algorithm into an off-the-shelf platform with limited resources to deploy this controller into a distributed system by approximating the Newton-Raphson solution into a closed form $\mathcal{O}(1)$ expression. We also plan to extend this controller to perform online learning of an evolving system's kinematics, further building up on our open-source {nn4mc} framework \cite{nn4mc}.

\vspace{-10pt}

\begin{acknowledgement}
We would like to thank Dr. Don Soloway, Prof. Bradley Hayes and  CAIRO Lab at CU Boulder and Cooper Simpson. This research has been supported by the Air Force Office of Scientific Research (Grant No. 83875-11094), we are grateful for this support.
\end{acknowledgement}

\vspace{-30pt}

%
%
%
\bibliographystyle{spphys}

\end{document}